\documentclass{article} % For LaTeX2e
\usepackage[final]{colm2026_conference}

\usepackage{microtype}
\usepackage{hyperref}
\usepackage{url}
\usepackage{booktabs}
\usepackage{graphicx}
\usepackage{subcaption}
\usepackage{caption}
\usepackage{amsmath}
\usepackage{amssymb}
\usepackage{listings}
\usepackage{xcolor}
\usepackage{placeins} % \FloatBarrier keeps each section's figures from
\usepackage{float}    % [H] pins each figure immediately after the
\definecolor{darkblue}{rgb}{0, 0, 0.5}
\hypersetup{colorlinks=true, citecolor=darkblue, linkcolor=darkblue, urlcolor=darkblue}

\lstdefinestyle{pycode}{
  language=Python,
  basicstyle=\ttfamily\small,
  keywordstyle=\bfseries,
  commentstyle=\itshape,
  showstringspaces=false,
  breaklines=true,
  columns=fullflexible,
  frame=single,
  numbers=left,
  numberstyle=\tiny,
  xleftmargin=1.5em,
}

\title{Why Does Robustness Reduce Superposition?}

\author{Adam Elimadi \\
Independent Researcher\\
\texttt{elimadadam@gmail.com}}

\begin{document}

\maketitle
% Override the running head set by \maketitle so it reflects the
% workshop acceptance rather than the generic "final" COLM string.
\lhead{Accepted at the COLM 2026 Workshop on AI Interpretability (AIW)}

\begin{abstract}
The study of adversarial examples and their origins remains an open area of research. Mechanistic interpretability, and superposition in particular, offers new avenues for approaching this problem. \citet{gorton2025adversarial} demonstrate that adversarial examples arise from superposition and show empirically that adversarial training reduces superposition, yet provide no mechanistic account of why this occurs. We present an empirical explanation inspired by the feature taxonomy of \citet{ilyas2019adversarial}, tracing the following chain of causalities: adversarial training abandons non-robust features $\rightarrow$ fewer total features to represent $\rightarrow$ less superposition.
\end{abstract}

\section{Introduction}
\label{sec:intro}

Explaining the existence of adversarial examples has been an open problem since \citet{goodfellow2015explaining} first showed that imperceptible perturbations reliably fool otherwise accurate models. \citet{ilyas2019adversarial} offered an influential account: models rely on non-robust features, patterns that are predictive on clean data but do not survive small perturbations. This explains classification behavior, but says nothing about how those features are represented internally. \citet{gorton2025adversarial} supplied that missing piece by connecting adversarial vulnerability to superposition --- the packing of more features into a representation than it has dimensions to hold cleanly \citep{elhage2022toy} --- showing that adversarial examples exploit superposed representations, and that adversarial training empirically reduces superposition. What their result does not explain is the mechanism: why training against perturbations should change how many features a model chooses to represent at all.

This paper supplies that mechanism. Using the same toy-model setup as \citet{gorton2025adversarial}, we show that adversarially trained models represent systematically fewer features than standardly trained ones, and that the discarded features are exactly the ones \citet{ilyas2019adversarial}'s taxonomy would classify as non-robust. We establish this first indirectly, through interference geometry and represented feature count, and then directly, using a synthetic dataset in which each feature's robust/non-robust identity is assigned in advance, rather than inferred from model behavior after training.

Our contributions are three-fold:
\begin{itemize}
    \item We show that adversarially trained models consistently drop more features than standardly trained models, across our sparsity sweep.
    \item We show that dropped features carry higher average interference than retained features, and that retained features cluster near antipodal interference ($\approx -1$), the configuration that minimizes cross-feature interference.
    \item Using a synthetic dataset with a known ground-truth partition, we show that the dropped features correspond exactly to the non-robust feature set.
\end{itemize}

\section{Background}
\label{sec:background}

\subsection{Superposition}
\label{sec:superposition}

Superposition occurs when a model encodes more features than there are neurons \citep{elhage2022toy}. We control it by intervening on sparsity when training toy models, following the setup of \citet{elhage2022toy}. Sparsity is governed by a threshold $S \in [0,1]$, which determines the activity of each component of the input $x \in \mathbb{R}^m$:
\[
[x]_i =
\begin{cases}
x_i & \text{if } x_i > S \\
0 & \text{otherwise}
\end{cases}
\]
A feature $x_i$ is therefore active with probability $p(A) = 1 - S$.

\subsection{Interference}

\citet{elhage2022toy} define interference as the geometry of features in activation space, measured by the degree of non-orthogonality between feature directions. For two feature directions $W_{:,i}, W_{:,j} \in \mathbb{R}^n$, they are orthogonal when $W_{:,i}^\top W_{:,j} = 0$, and exhibit interference when $W_{:,i}^\top W_{:,j} \neq 0$. The more strictly positive this inner product, the more likely spurious activations are to corrupt the reconstruction of feature $i$, constituting harmful interference.

\subsection{Adversarial Examples}

An adversarial example is a perturbation of $x \in \mathbb{R}^m$, imperceptible to humans, that maximizes the MSE loss within an $\ell_2$ ball of radius $\varepsilon$:
\begin{equation}
x_{\text{adv}} = x + \operatorname*{arg\,max}_{\|\delta\|_2 \le \varepsilon} \mathcal{L}(x + \varepsilon \cdot \nabla_x \mathcal{L})
\label{eq:adv}
\end{equation}
Following \citet{gorton2025adversarial}, a small noise term is added to $x$ prior to attack generation to avoid gradient masking, after which a one-step $\ell_2$ gradient attack is applied. The concept builds on foundational work in adversarial robustness \citep{goodfellow2015explaining}.

\FloatBarrier
\section{Setup}

\subsection{Toy Models}

We adopt the simplified toy model of \citet{elhage2022toy} as employed by \citet{gorton2025adversarial}. Let $W \in \mathbb{R}^{n \times m}$, with $n = 20$ and $m = 100$. Given inputs $x \sim \mathcal{D}$, $\mathcal{D} \subset \mathbb{R}^m$, hidden representations are computed as $h = Wx \in \mathbb{R}^n$ and features are reconstructed via $\hat{x} = \text{ReLU}(W^\top h + b)$. The training objective is $\mathcal{L} = \|x - \hat{x}\|_2^2$. Higher sparsity leads the model to represent more features and, consequently, to exhibit more superposition.

\subsection{Measuring Superposition}

\citet{elhage2022toy} quantify superposition as $n / \|W\|_F$, the number of dimensions per feature. We adapt this by taking the squared Frobenius norm per dimension, obtaining a quantity proportional to the total energy distributed across hidden dimensions:
\begin{equation}
\Phi = \|W\|_F^2 / n
\label{eq:phi}
\end{equation}
$\Phi$ increases monotonically with superposition and sparsity.

\subsection{Representational Power}

We define representational power as the squared Frobenius norm of $W$, measuring the total energy distributed across all feature directions $W_{:,i}$:
\begin{equation}
P = \|W\|_F^2
\label{eq:power}
\end{equation}
Note that $\Phi = P/n$, so representational power and superposition are proportional given fixed $n$.

\subsection{Feature Representation Threshold}

To determine which features are represented and which are dropped, we measure the norm $\|W_{:,i}\|_2$ of each feature column following \citet{elhage2022toy}, applying the threshold $\tau = 0.01$:
\begin{equation}
\|W_{:,i}\|_2
\begin{cases}
< \tau \Rightarrow \text{feature } i \text{ is dropped} \\
\ge \tau \Rightarrow \text{feature } i \text{ is represented}
\end{cases}
\label{eq:threshold}
\end{equation}

\subsection{Measuring Interference}

Let $G = WW^\top \in \mathbb{R}^{n \times n}$ be the Gram matrix of $W$. The interference matrix is obtained by zeroing the diagonal, retaining only pairwise cross-feature terms:
\begin{equation}
I = G - \text{diag}(G)
\label{eq:interference-matrix}
\end{equation}
The per-feature total interference is obtained by summing $I$ along each row:
\begin{equation}
\iota_i = \sum_{j=1}^{n} I_{ij}
\label{eq:iota}
\end{equation}
Given the binary mask induced by Eq.~\eqref{eq:threshold}, the mean total interference of kept and dropped features is then computed separately as:
\begin{equation}
\bar{\iota}_{\text{kept}} = \frac{1}{|K|} \sum_{i: m_i = 1} \iota_i, \qquad
\bar{\iota}_{\text{dropped}} = \frac{1}{|D|} \sum_{i: m_i = 0} \iota_i
\label{eq:iota-bar}
\end{equation}

\subsection{Adversarial Training Protocol}

Following \citet{gorton2025adversarial} and the adversarial training framework of \citet{madry2019towards}, models are trained over a sweep of sparsity levels on a mixture of clean and adversarial examples:
\begin{equation}
\mathcal{L}_{\text{adv}} = \alpha \cdot \mathcal{L}(x) + (1 - \alpha) \cdot \mathcal{L}(x_{\text{adv}})
\label{eq:adv-loss}
\end{equation}
We set $\alpha = 0.5$ to balance clean and robust accuracy and prevent collapse in either regime. Attacks are generated via Eq.~\eqref{eq:adv} with $\varepsilon = \frac{0.1}{B}\sum_{x \in \mathcal{B}} \|x\|_2$, where $\mathcal{B}$ denotes the training batch of size $B$. Models are trained for 150{,}000 steps at a learning rate of $10^{-3}$.

\subsection{Sparsity Sweep}

To cleanly isolate our theory empirically, we exclude both extremes of the sparsity range. At $S \le 0.7$, models drop most features regardless of training regime, as sparsity is too low to motivate superposition and interference is easily avoided. At $S \ge 0.98$, both models represent all features at negligible interference cost. We focus on the $\ell_2$ threat model, a standard adversarial threat model studied extensively in prior work \citep{carlini2017towards}. We therefore sweep over $\mathcal{S} = \{0.78, 0.80, 0.88, 0.90\}$, the range in which the adversarially trained model consistently drops more features than the standard model --- an effect we attribute to the presence of distinct feature classes, rather than to the routine optimization of feature benefit against interference cost.

\FloatBarrier
\section{Representational Power and Feature Dropping}
\label{sec:repr-power}

We train two models over $\mathcal{S}$ --- one under the adversarial training protocol (Eq.~\eqref{eq:adv-loss}) and one under standard training --- with $x \sim \mathcal{U}([0,1]^m)$.

\subsection{Replication}

We first reproduce the core finding of \citet{gorton2025adversarial} within our experimental range before proceeding to explain it. Measuring $\Phi$ (Eq.~\eqref{eq:phi}) under both training regimes across $\mathcal{S}$, we observe that adversarially trained models consistently exhibit lower superposition than their standardly trained counterparts, with the gap widening as sparsity increases.

\subsection{Representational Power and Feature Dropping}

We measure representational power (Eq.~\eqref{eq:power}) across $\mathcal{S}$ for each training regime, and, to investigate why the adversarially trained model exhibits lower representational power, we compute $\|W_{:,i}\|_2$ for all $i$. Applying Eq.~\eqref{eq:threshold}, we find that adversarially trained models drop more features than standardly trained models across all $S \in \mathcal{S}$.

\begin{figure}[htbp]
    \centering
    \begin{subfigure}[b]{0.48\linewidth}
        \centering
        \includegraphics[width=\linewidth]{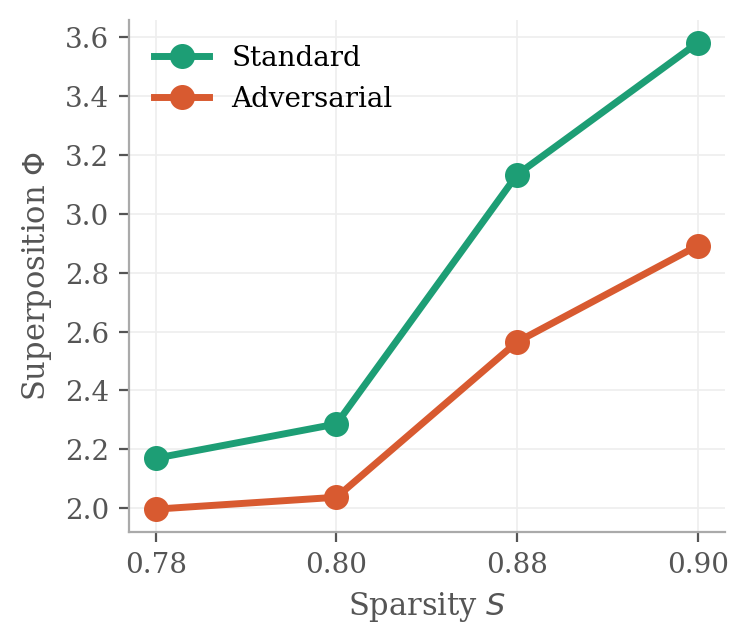}
        \caption{Superposition $\Phi$ (Eq.~\eqref{eq:phi}) across $\mathcal{S}$ for both training regimes. Adversarial training consistently reduces superposition, reproducing the finding of \citet{gorton2025adversarial} within our sparsity range.}
        \label{fig:superposition}
    \end{subfigure}
    \hfill
    \begin{subfigure}[b]{0.48\linewidth}
        \centering
        \includegraphics[width=\linewidth]{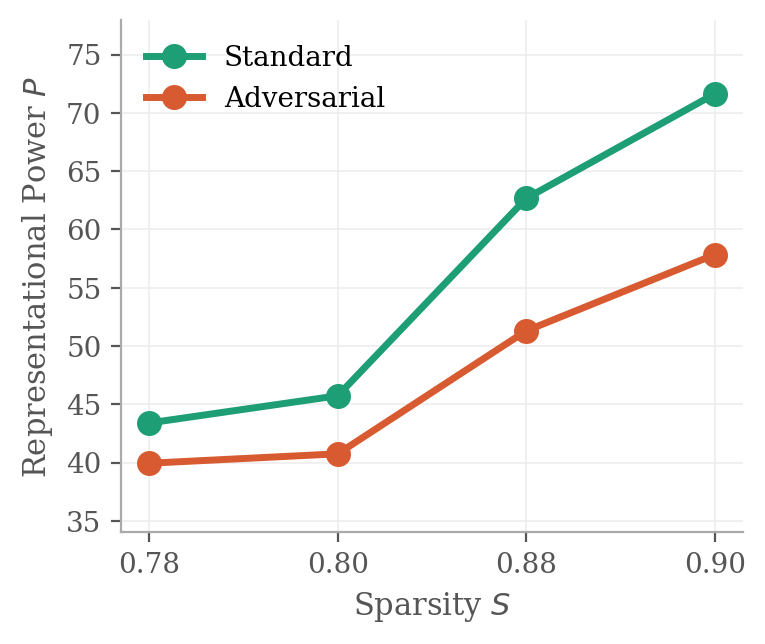}
        \caption{Representational power $P$ (Eq.~\eqref{eq:power}) across $\mathcal{S}$ for both training regimes. The adversarially trained model consistently exhibits lower representational power, with the gap widening as sparsity increases.}
        \label{fig:power}
    \end{subfigure}
    \caption{Superposition and representational power fall together as adversarial training discards features.}
    \label{fig:superposition-power}
\end{figure}
\FloatBarrier

\subsection{Feature Dropping and Interference}

We examine the dropped features through the lens of interference. Normalizing $W$ and computing the interference matrix (Eq.~\eqref{eq:interference-matrix}), we calculate each column's total interference with the rest, then average across retained and dropped groups separately. We find that dropped features either exhibit higher interference than retained ones, or --- when both groups are negative --- the adversarial model preferentially retains those closest to $-1$.

\begin{figure}[htbp]
    \centering
    \begin{subfigure}[b]{0.48\linewidth}
        \centering
        \includegraphics[width=\linewidth]{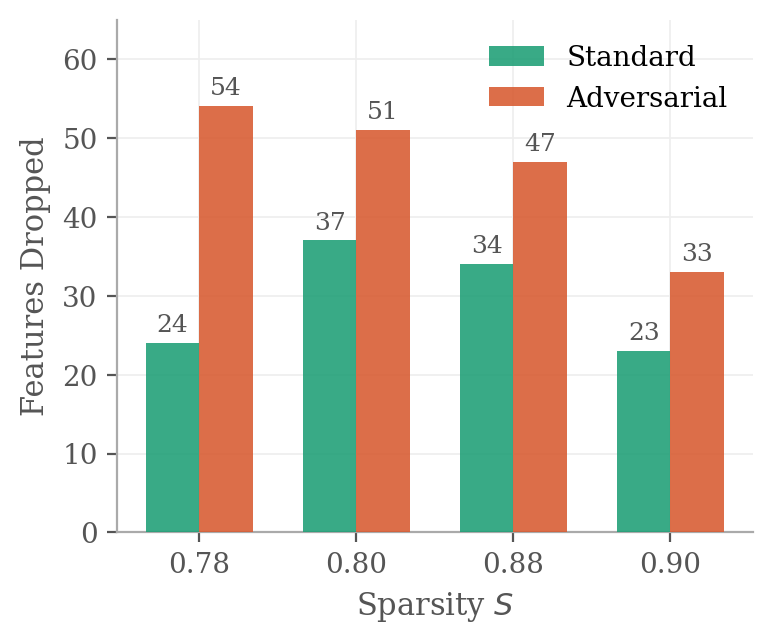}
        \caption{Number of features dropped at each $S \in \mathcal{S}$ under adversarial and standard training. The adversarially trained model drops more features at every sparsity level.}
        \label{fig:dropped}
    \end{subfigure}
    \hfill
    \begin{subfigure}[b]{0.48\linewidth}
        \centering
        \includegraphics[width=\linewidth]{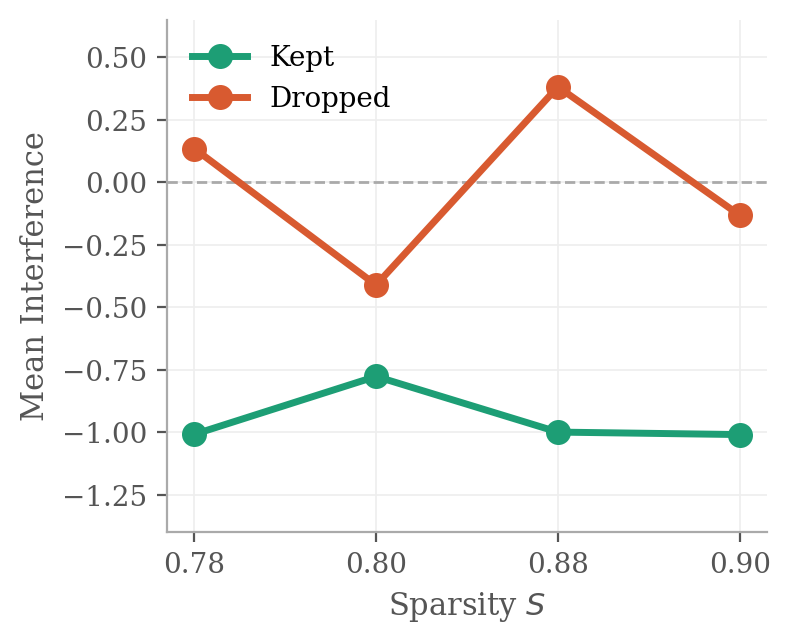}
        \caption{Mean interference of kept versus dropped features across $\mathcal{S}$ for the adversarially trained model. Dropped features consistently exhibit higher or less-negative mean interference than retained features.}
        \label{fig:interference}
    \end{subfigure}
    \caption{Adversarial training drops more features than standard training, and the dropped features are the ones with the most costly interference geometry.}
    \label{fig:dropped-interference}
\end{figure}
\FloatBarrier

These results establish that adversarially trained models reduce superposition by dropping more features than their standardly trained counterparts, and that the dropped features are those the model identifies as geometrically costly. The preference for retaining features with interference nearest to $-1$ is particularly telling: a value of $-1$ corresponds to antipodal feature directions, the configuration that maximally suppresses cross-feature activation. This aligns with \citet{elhage2022toy}, who observe that models preferentially pack features into opposing directions to minimize spurious reconstruction. A key question remains: do the dropped features correspond to the robust or non-robust classes of \citet{ilyas2019adversarial} and \citet{li2025adversarial}?

\FloatBarrier
\section{Robust and Non-Robust Feature Recovery}

\subsection{Structured Data Construction}
\label{sec:structured-data}

To answer this question, Section~\ref{sec:repr-power} must be reproduced under a data distribution where the ground-truth identity of robust and non-robust features is known a priori. We therefore construct a synthetic dataset that operationalizes the feature taxonomy of \citet{ilyas2019adversarial} and \citet{li2025adversarial} under the constraints of our toy model setting.

Robust features are broadly characterized as high-signal directions that encode stable, generalizable patterns, whereas non-robust features are low-amplitude, high-frequency directions that are predictive under clean inputs but easily disrupted by adversarial perturbations. More precisely, \citet{ilyas2019adversarial} show that models depend substantially on non-robust features for classification, yet those features cease to correlate with the correct label after an attack. \citet{li2025adversarial} further characterize this taxonomy by establishing that robust features carry greater signal amplitude than non-robust ones, while non-robust features are denser across the feature space.

We note, however, that our setting differs substantially from the classification context in which this taxonomy was originally defined. Our experiments are conducted on toy models optimizing a reconstruction loss (MSE), which bears structural similarities to linear regression but falls well short of the complexity of classification over real-world data with semantic labels. The instantiation of robust and non-robust features below is therefore necessarily an approximation made under these constraints.

We instantiate this taxonomy as follows. Let $x \sim \mathcal{U}([0,1]^m)$ with the sparsity mask of Section~\ref{sec:superposition} applied. A uniformly random permutation $\pi$ of $\{1, \ldots, m\}$ partitions the feature indices into a robust set $\mathcal{R} = \{\pi(1), \ldots, \pi(n_r)\}$ and a non-robust set $\mathcal{N} = \{\pi(n_r + 1), \ldots, \pi(m)\}$, with $|\mathcal{R}| = n_r$ and $|\mathcal{N}| = m - n_r$. Each partition is then amplitude-scaled according to:
\begin{equation}
x_{:,i} \leftarrow
\begin{cases}
a_r \cdot x_{:,i} & i \in \mathcal{R} \\
a_{nr} \cdot x_{:,i} & i \in \mathcal{N}
\end{cases}
\label{eq:amplitude}
\end{equation}
with $a_r = 6.0$ and $a_{nr} = 0.2$. Robust features thus carry high amplitude --- strong, stable signal --- while non-robust features carry low amplitude, making them more susceptible to erasure under $\ell_2$ perturbations. The partition $(\mathcal{R}, \mathcal{N})$ is retained at generation time and serves as the ground truth against which the model's dropped feature indices are evaluated. The full data generation procedure is given in Appendix~\ref{app:code}.

To remain consistent with the definition of \citet{li2025adversarial}, we set $|\mathcal{N}| > |\mathcal{R}|$, making non-robust features denser than robust ones. We note that models can be sensitive to the amplitude ratio between robust and non-robust features; the specific values ($a_r = 6.0$, $a_{nr} = 0.2$) were chosen to reflect the signal-to-noise distinction while maintaining numerical stability during training.

\subsection{Results}

Training on data generated via Eq.~\eqref{eq:amplitude}, we obtain the same findings as in Section~\ref{sec:repr-power}. The structured data, however, yields a substantially cleaner pattern: the adversarially trained model drops exactly $|\mathcal{N}| = 70$ features at every $S \in \mathcal{S}$, regardless of sparsity level. Notably, at $S \in \{0.88, 0.90\}$, the standardly trained model drops no features at all --- yet the adversarially trained model still drops exactly $|\mathcal{N}|$ --- indicating that, despite the model having enough sparsity to represent all features, adversarial training remains highly sensitive to non-robust features. This sensitivity being sparsity-invariant suggests that those features directly interfere with the model's ultimate goal:
\[
\min_\theta \; \mathbb{E}_{x \sim \mathcal{D}}[\mathcal{L}(x_{\text{adv}}; \theta)]
\]

\begin{figure}[htbp]
    \centering
    \includegraphics[width=0.95\linewidth]{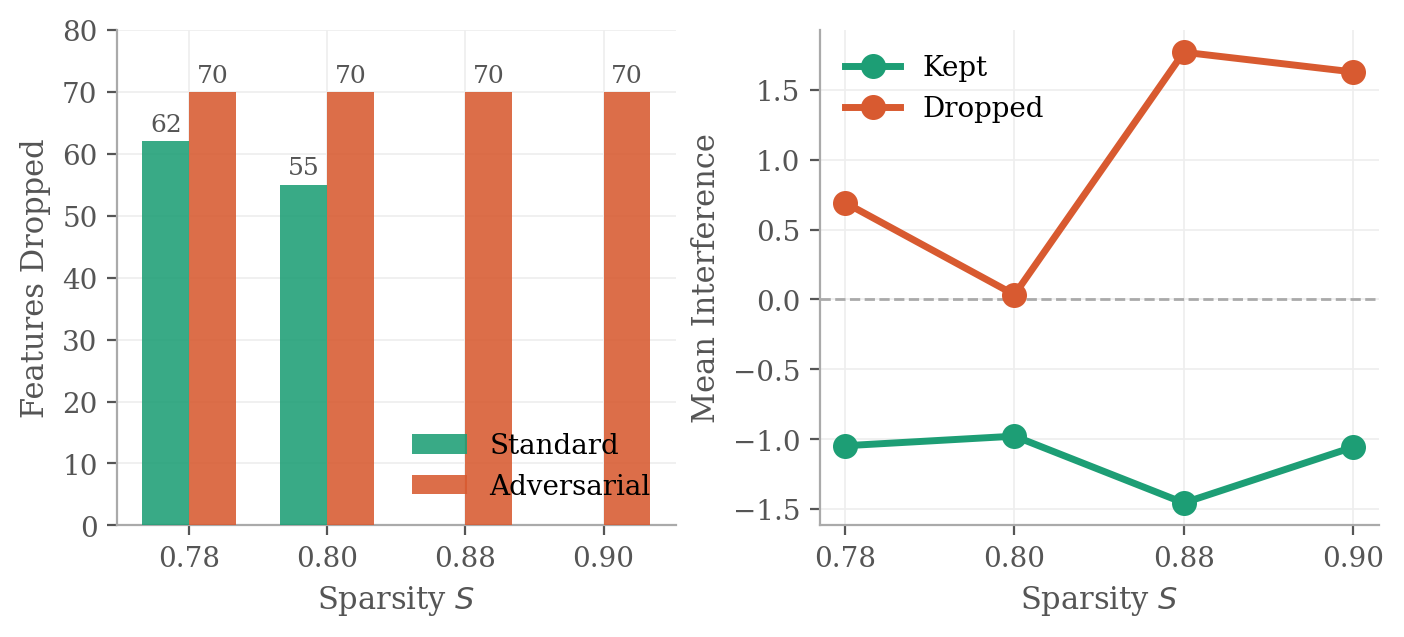}
    \caption{\textbf{Left:} Features dropped at each $S \in \mathcal{S}$ under structured data. The adversarially trained model drops exactly $|\mathcal{N}| = 70$ at every sparsity level. \textbf{Right:} Mean interference of kept versus dropped features across $\mathcal{S}$ for the adversarially trained model under structured data. Dropped features carry positive net interference; kept features carry negative net interference.}
    \label{fig:structured-data}
\end{figure}
\FloatBarrier

In order to determine whether the dropped features are primarily the non-robust ones, we map the ground truth indices of non-robust features retrieved at data creation time and compare them to the indices of the dropped features retrieved at the end of training. We find that they correspond exactly to $\mathcal{N}$ at every $S \in \mathcal{S}$, confirming that the adversarial model explicitly targets the non-robust feature set.

\begin{center}
    \includegraphics[width=\linewidth]{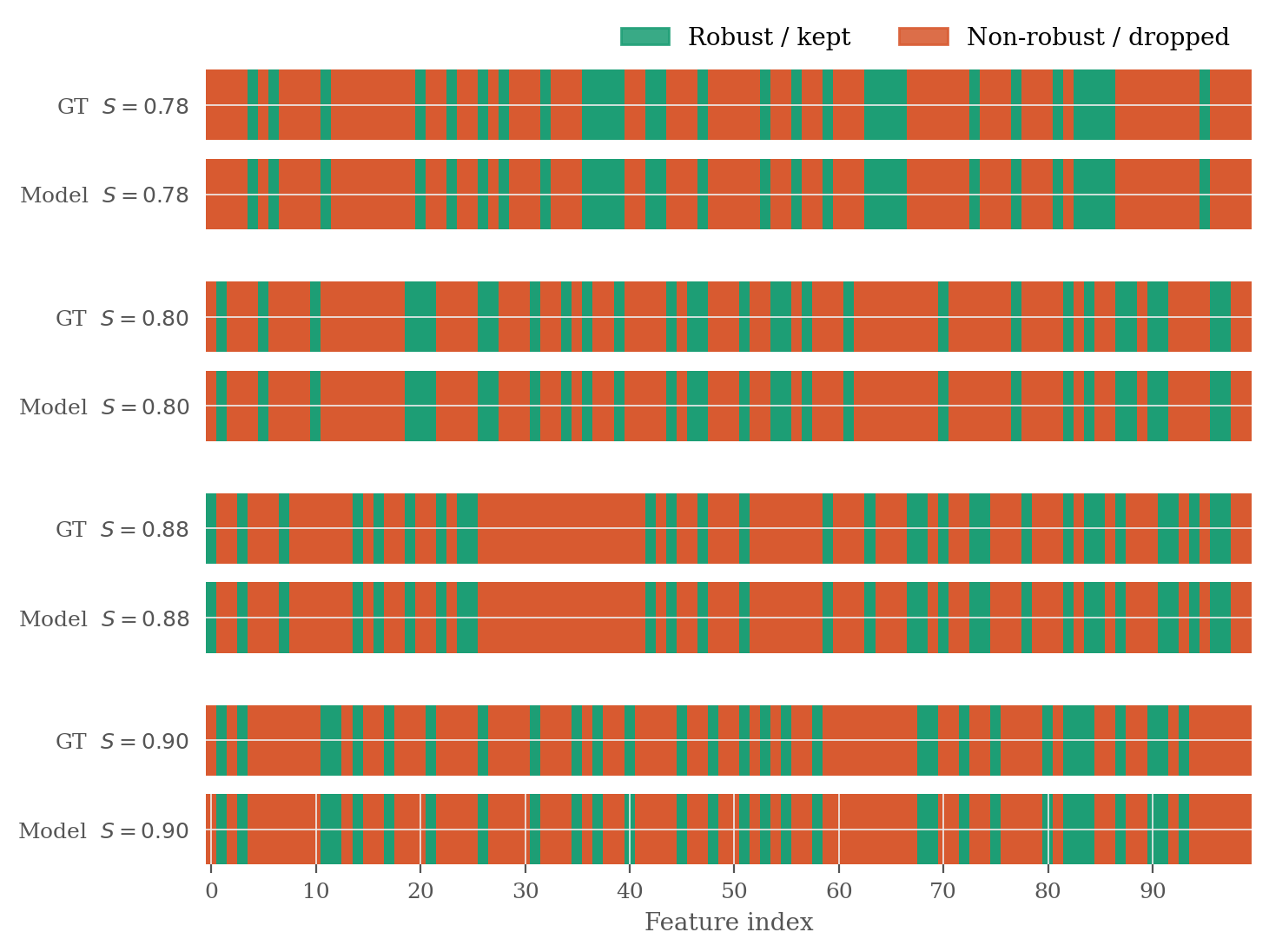}
    \captionof{figure}{Ground truth partition $(\mathcal{R}, \mathcal{N})$ versus model-dropped indices across $\mathcal{S}$. For each sparsity level, the top row shows the ground truth and the bottom row shows the adversarial model output. Identical patterns confirm that the adversarial model precisely targets $\mathcal{N}$ at every sparsity level.}
    \label{fig:ground-truth}
\end{center}
\FloatBarrier

\FloatBarrier
\subsection{Theoretical Intuition}

Our empirical findings raise a natural question: why do adversarially trained models systematically drop non-robust features? We offer a theoretical explanation grounded in the properties of non-robust features under adversarial attack. The argument is conceptual rather than a formal derivation, but it provides intuition for the observed behavior.

Let $\mathcal{D} = \{(x, y)\}$ with $y = x$, parameters $\theta$, loss $\mathcal{L}$, and adversarial examples $x_{\text{adv}}$ generated via an $\ell_2$-bounded attack. We conceptualize adversarial training's primary objective as:
\begin{equation}
\min_\theta \; \mathbb{E}_{(x,y) \sim \mathcal{D}}[\mathcal{L}(x_{\text{adv}}, y; \theta)]
\label{eq:objective}
\end{equation}
Building on \citet{ilyas2019adversarial}, we define a feature $i$ as non-robust if it is highly sensitive to adversarial perturbations such that $x_{\text{adv},i} \not\approx x_i$, breaking the correspondence with its target $y_i = x_i$. Denote the set of non-robust features $\mathcal{N} \subset [m]$.

We hypothesize that features in $\mathcal{N}$ interfere with the adversarial training objective via two complementary mechanisms:

\paragraph{Direct loss inflation.} Since $\mathcal{L} = \|y - \hat{x}\|_2^2 = \sum_i (y_i - \hat{x}_i)^2$ and $y$ is fixed under attack, each flipped feature $i \in \mathcal{N}$ contributes a non-zero penalty term to the loss. Even a single flipped feature increases $\mathcal{L}$, and the effect compounds: inputs with more non-robust features incur proportionally higher loss. Let $k$ denote the number of non-robust features flipped per adversarial example. This gives:
\begin{equation}
\mathcal{L}_{\text{adv}}(x_{\text{adv}}, k{=}0) < \mathcal{L}_{\text{adv}}(x_{\text{adv}}, k{=}1) < \cdots < \mathcal{L}_{\text{adv}}(x_{\text{adv}}, k{=}|\mathcal{N}|)
\label{eq:loss-inflation}
\end{equation}
Dropping all features in $\mathcal{N}$ eliminates these penalty contributions entirely and gives $P(\mathcal{N}_{\text{flipped}} \mid x_{\text{adv}}) = 0$, thus enabling the model to minimize Eq.~\eqref{eq:adv-loss} freely.

\paragraph{Costly interference.} Non-robust features exhibit costly feature geometry, as evidenced by their mean total interference lying in the range $[0, 2)$ compared to retained features clustering near $-1$. This costly geometry induces spurious cross-feature activations that corrupt the reconstruction and further inflate $\mathcal{L}$.

Together, these two mechanisms explain why adversarially trained models prune non-robust features: they are the primary obstacle to minimizing the adversarial objective, and removing them reduces superposition while improving robustness.

\FloatBarrier
\section{Discussion}

We have established a clean causal chain explaining why adversarial training reduces superposition: adversarially trained models drop more features than standard models, leaving fewer features to encode in the same dimensional space, and thus reducing superposition. We further demonstrated that the dropped features correspond precisely to non-robust features as defined by \citet{ilyas2019adversarial}.

These results answer our core question but raise new ones. First, why do adversarially trained models preserve the geometry of remaining features, keeping interference values largely intact \citep{gorton2025adversarial}? Second, why does adversarial training preferentially align features in opposite directions, enabling antipodal superposition? Understanding these emergent behaviors would deepen our mechanistic account of adversarial robustness.

Our findings remain constrained by the toy model setting. The natural next step is validating our theory on real-world models using mechanistic tools like sparse autoencoders (SAEs), moving beyond controlled toy settings.

Recent work at the intersection of robustness and interpretability suggests a promising research direction. We are particularly interested in two questions: First, does superposition reduction in real adversarial models translate to meaningfully improved interpretability? Second, can we intentionally leverage the robust/non-robust feature trade-off to reduce polysemanticity in a controlled manner, without sacrificing critical representations?

\section*{Acknowledgments}
Claude (Anthropic) was used to assist with \LaTeX{} formatting and editorial refinements. All scientific content, experimental design, results, and writing are the author's own.

\bibliography{main}
\bibliographystyle{colm2026_conference}

\appendix

\section{Data Generation Code}
\label{app:code}

The full procedure used to construct the structured robust/non-robust dataset of Section~\ref{sec:structured-data} is given below.

\begin{lstlisting}[style=pycode, caption={Structured data generation with ground-truth robust/non-robust partition.}, label={lst:datagen}]
def create_data(sparsity, num_samples, n_features=100, n_robust=30):
    values = torch.rand(num_samples, n_features)
    mask = values > sparsity
    x = values * mask.float()
    pi = torch.randperm(n_features)
    robust_idx = pi[:n_robust]
    non_robust_idx = pi[n_robust:]
    x[:, robust_idx] *= 6.0
    x[:, non_robust_idx] *= 0.2
    return x, x.clone(), robust_idx, non_robust_idx
\end{lstlisting}

\end{document}